\documentclass{article}

\usepackage[dblblindworkshop,final]{neurips_2026}
\workshoptitle{VLM4RWD}

\usepackage[utf8]{inputenc}
\usepackage[T1]{fontenc}
\usepackage{hyperref}
\usepackage{url}
\usepackage{booktabs}
\usepackage{amsfonts}
\usepackage{amsmath}
\usepackage{nicefrac}
\usepackage{microtype}
\usepackage{graphicx}
\usepackage{xcolor}
\usepackage{makecell}
\usepackage{float}
\usepackage{graphicx}
\usepackage{listings}
\usepackage{xcolor}
\usepackage{float} 

\title{Grounded Exploration for Open-Ended Image Editing}

\author{
\textbf{Jieyu Zhou}$^{1,2}$ \quad
\textbf{Mehdi Ben Ayed}$^{2}$ \quad
\textbf{Shubham Gupta}$^{2}$ \quad
\textbf{Zhenyu Tan}$^{2}$ \\
\textbf{Jiajing Xu}$^{2}$ \quad
\textbf{Dmitry Kislyuk}$^{2}$ \quad
\textbf{Eric Tzeng}$^{2}$ \\
\textbf{Christopher J. MacLellan}$^{1}$ \quad
\textbf{Charles Rosenberg}$^{2}$ \\[6pt]
$^{1}$Georgia Institute of Technology \quad
$^{2}$Pinterest \\[4pt]
\texttt{\{jzhou625,cmaclell\}@gatech.edu} \\
\texttt{\{mbenayed,shubhamgupta,zhenyutan,jiajing,} \\
\texttt{dkislyuk,etzeng,crosenberg\}@pinterest.com}
}

\begin{document}

\maketitle

\begin{abstract}
Open-ended image-editing requests, such as ``decorate this room,'' admit multiple valid realizations. Direct sampling may produce ungrounded variation, whereas explicit diversity prompting often produces superficial differences. We study how to ground these requests without prematurely eliminating their legitimate ambiguity. We introduce a two-stage framework  that first identifies diverse, feasible edit scopes and then realizes them as executable edit instructions. Across 300 open-ended image-editing tasks, our approach achieves 24.6\% higher mean semantic dispersion and 17.1\% higher mean groundedness than direct diversity prompting, while producing more valid and distinct realizations. These results show that structuring feasible edit directions before realization supports more useful grounded exploration.
\end{abstract}

\section{Introduction}
Image generation and editing systems are often evaluated on requests with relatively specific targets \cite{xiang2025promptsculptor, zhang2023magicbrush,sheynin2024emu}, such as removing an object or changing its color. Prior image-editing systems therefore typically optimize for a single best edit for a given instruction \cite{zhao2026qwen, xu2026edit}. Yet many real-world creative requests \cite{collins2026understanding} are fundamentally more open-ended: users may instead ask for “ideas to make my bedroom feel cozier” or “ways to improve this space.” Such requests do not specify a single desired outcome, but admit many plausible realizations. Supporting them therefore requires more than producing a valid edit—it requires exploring the space of possible solutions that are consistent with the user's intent. In practice, however, because each generation conditions on the same request and source image, repeated sampling often remains concentrated around the same high-level interpretation. For example, a model may respond to “make this room cozier” with several variations that all add plants, while overlooking qualitatively different directions such as introducing rugs, wall art, warmer lighting, shelving, or combinations of these changes. Conversely, explicitly prompting for diversity can increase variation without ensuring that the alternatives correspond to meaningful or plausible editing directions.

This exploration problem is especially challenging for image editing, where the solution space is constrained by an existing source image. Unlike unconstrained image generation, a successful edit must not only reflect the user's underspecified intent, but also remain grounded in the visible objects, spatial layout, scene structure, and implicit preservation constraints of the input. The model must therefore infer where the request leaves room for change and where the source image constrains it, while still exploring substantially different ways of satisfying the request. These objectives can be in tension: staying too close to the most obvious interpretation leads to redundant alternatives, whereas aggressively seeking diversity can introduce edits that alter unrelated regions, violate spatial constraints, or drift away from the user's intent. We therefore view open-ended image editing as a problem of grounded exploration: generating multiple meaningfully different editing directions while respecting the solution space supported by both the request and the source image.

To move beyond prior work of producing a single best edited image, we separate exploration from visual execution. We designed a two-stage workflow: it first constructs an input-adaptive scaffold that captures relevant visual constraints while preserving genuinely open decisions, and then uses this scaffold to generate multiple semantically distinct, executable edit realizations. Across 300 open-ended image-editing tasks, our approach achieves 24.6\% higher mean semantic dispersion and 17.1\% higher mean groundedness than direct diversity prompting, while also producing more valid and distinct realizations. These results suggest that effective exploration benefits from explicitly structuring high-level edit directions before realization, rather than simply prompting for diversity.

\section{Grounded Exploration for Open-Ended Image Editing}
Given a source image and an underspecified request, our goal is to generate a set of $K$ meaningfully different candidate edits while remaining grounded in the user's intent and the visible scene. Rather than relying on repeated sampling from an image editing model to discover alternatives implicitly, our method separates spatial exploration from concrete realization.

\paragraph{Stage 1: Exploring spatial edit scopes.}
The first stage generates K distinct spatial scopes as natural-language descriptions that are compatible with the image and request. Each scope specifies the primary editable region, the permitted extent of change, and a preservation boundary for the rest of the image. For example, a request to give a room a more designer look may be applied to the sofa area, the fireplace, the workspace, or the entire seating area. At this stage, the model does not choose a particular style, material, or object replacement. It determines only where an edit may occur and how much of the visible scene it may affect.

\paragraph{Stage 2: Realizing concrete edits.}
The second stage generates one natural-language edit plan for each spatial scope. Each plan is a self-contained description of the specific objects, materials, and visual treatments to modify while respecting the corresponding editable region and preservation boundary. For example, a fireplace scope may be realized by refinishing the surround, while a seating-area scope may coordinate changes to the sofa, rug, and coffee table. The resulting K plans are passed directly to the same image editing model and executed independently.

\begin{figure}
    \centering
    \includegraphics[width=\linewidth]{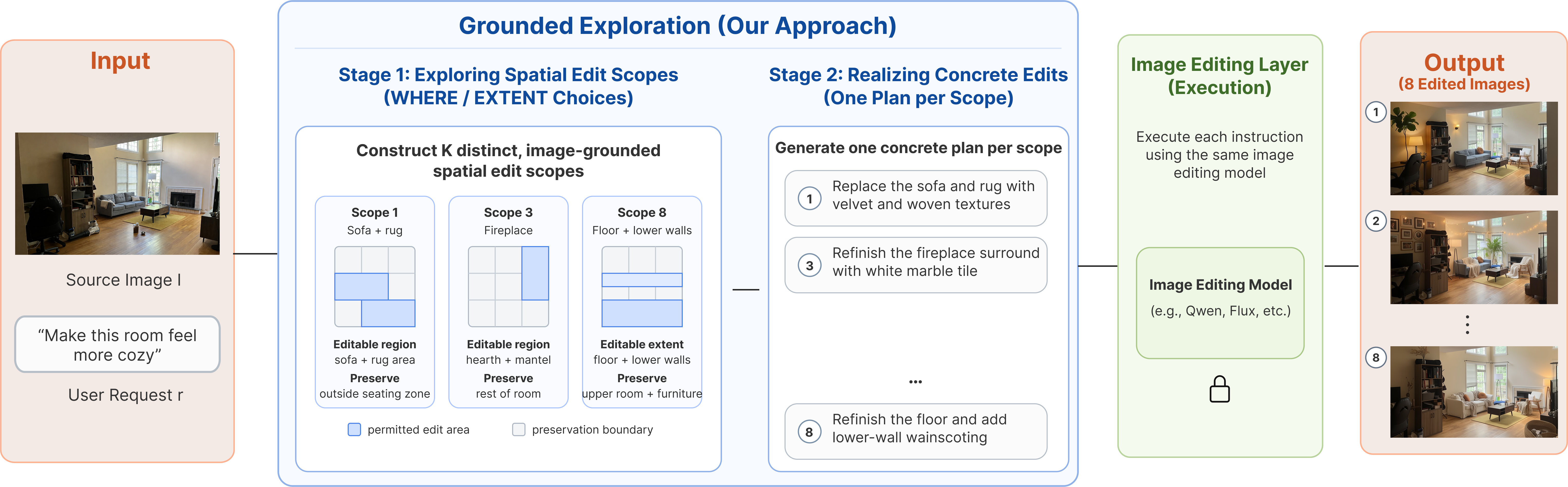}
    \caption{Overview of our two-stage grounded exploration framework. Stage 1 constructs image-grounded edit scopes. Stage 2 generates concrete edit instructions.}
    \label{fig:placeholder}
\end{figure}

\section{Experiments}

We evaluate whether explicitly structuring exploration produces candidate sets that are both more diverse and more grounded than alternatives based on repeated sampling or generic diversity prompting. Our evaluation uses 300 open-ended image-editing tasks sampled from an internal evaluation set collected during employee testing of an online design platform. Each user request generated eight candidates.
\subsection{Evaluation Metrics}

Our primary evaluation is set-level, since the goal is to produce a useful collection of grounded alternatives rather than identify a single best edit. We use GPT-5.5 as a VLM judge, providing the source image, original user request, and eight generated candidates for each method. We report two \textbf{1--5 set-level ratings}: \textbf{Instruction Following}, which measures how well the generated set satisfies the requested editing intent and scope, and \textbf{Groundedness}, which measures how well the outputs preserve the identity, structure, spatial relationships, and unrelated content of the source image. We additionally report \textbf{Valid@8}, the number of candidates that both satisfy the request and remain sufficiently grounded in the source.

After fixing candidate validity in the first judge call, we use two additional calls to characterize diversity at different semantic levels. \textbf{Direction Coverage@8} counts the valid high-level editing directions represented in the set, where directions differ in target, function, layout, intervention strategy, or interpretation. Differences only in color, material, texture, accessory choice, or minor placement remain within the same direction. A separate call clusters valid candidates into concrete realizations. \textbf{Distinct Realizations@8} counts these clusters, allowing multiple realizations to belong to the same broader direction. Finally, \textbf{Dispersion} averages pairwise semantic separation over all 28 candidate pairs: invalid or redundant pairs receive no credit, related realizations receive partial credit, and different high-level directions receive full credit.

\subsection{Experimental Setup}

Our experiments study where and how to introduce exploration so that the resulting candidate set is diverse while remaining grounded in the source image and user request. We ask two questions: (1) whether grounded diversity is better obtained through prompt-level exploration or through stochastic image-model sampling, and (2) how prompt-level exploration should be structured.

\textbf{Image-model sampling.} We provide the same source image and original user request to an image editing model and generate eight outputs by varying only the random seed. We evaluate three image backends: \texttt{Qwen/Qwen-Image-Edit-2509}, \texttt{gemini-2.5-flash-image}, and \texttt{black-forest-labs/FLUX.2-klein-4B}. This setting tests whether stochastic image generation alone can produce meaningfully different edits without generating alternative edit instructions.

\textbf{Instruction-space exploration.} All prompt-level methods use \texttt{Qwen/Qwen3.6-35B-A3B} to generate eight alternative edit instructions, which are then rendered using a fixed downstream backend, \texttt{Qwen/Qwen-Image-Edit-2509}. We consider two unstructured baselines. \emph{Naive} independently samples eight instructions, whereas \emph{Prompted Diversity} requests eight diverse instructions in a single call. As an ablation, \emph{Realization only} uses our Stage-2 prompt to jointly generate eight concrete realizations without the preceding exploration scaffold. We additionally evaluate \emph{High-level exploration}, an alternative two-stage strategy that first identifies a small set of image-grounded readings of the request, while using the same Stage-2 procedure. 

\subsection{Results}

Table~\ref{tab:main_results} summarizes the set-level evaluation. Overall, structured exploration improves candidate-set quality over generic diversity prompting.

\begin{table}[t]
\centering
\small
\resizebox{\columnwidth}{!}{%
\begin{tabular}{lcccccc}
\toprule
\textbf{Method} &
\makecell{\textbf{Instruction}\\\textbf{rating 1-5}} &
\makecell{\textbf{Groundedness}\\\textbf{rating 1-5}} &
\textbf{Valid@8} &
\makecell{\textbf{Direction}\\\textbf{Coverage@8}} &
\makecell{\textbf{Distinct}\\\textbf{Realizations@8}} &
\makecell{\textbf{Semantic}\\\textbf{Dispersion}} \\
\midrule

Realization only
& 3.77 $\pm$ 0.10
& 3.73 $\pm$ 0.09
& 5.83 $\pm$ 0.23
& \textbf{4.17 $\pm$ 0.21}
& 5.63 $\pm$ 0.23
& 0.504 $\pm$ 0.033 \\

High-level exploration
& 3.70 $\pm$ 0.09
& 3.77 $\pm$ 0.09
& 5.80 $\pm$ 0.23
& 4.16 $\pm$ 0.21
& 5.63 $\pm$ 0.23
& 0.504 $\pm$ 0.034 \\

Region-level exploration (ours)
& 3.74 $\pm$ 0.10
& 3.90 $\pm$ 0.08
& \textbf{6.05 $\pm$ 0.23}
& 4.00 $\pm$ 0.21
& \textbf{5.65 $\pm$ 0.22}
& \textbf{0.532 $\pm$ 0.033} \\

Prompted Diversity
& 3.57 $\pm$ 0.10
& 3.33 $\pm$ 0.10
& 5.33 $\pm$ 0.24
& 3.66 $\pm$ 0.20
& 5.20 $\pm$ 0.23
& 0.427 $\pm$ 0.034 \\

Naive
& 3.31 $\pm$ 0.09
& 3.38 $\pm$ 0.09
& 4.78 $\pm$ 0.22
& 2.90 $\pm$ 0.16
& 4.42 $\pm$ 0.20
& 0.319 $\pm$ 0.027 \\

Qwen
& 2.98 $\pm$ 0.10
& 3.26 $\pm$ 0.10
& 4.04 $\pm$ 0.24
& 2.46 $\pm$ 0.16
& 3.63 $\pm$ 0.22
& 0.242 $\pm$ 0.025 \\

Gemini
& \textbf{3.93 $\pm$ 0.10}
& 3.45 $\pm$ 0.10
& 5.74 $\pm$ 0.26
& 2.18 $\pm$ 0.12
& 4.56 $\pm$ 0.22
& 0.424 $\pm$ 0.032 \\

Flux
& 3.77 $\pm$ 0.14
& \textbf{3.95 $\pm$ 0.10}
& 5.71 $\pm$ 0.34
& 1.49 $\pm$ 0.13
& 2.85 $\pm$ 0.23
& 0.300 $\pm$ 0.032 \\

\bottomrule
\end{tabular}%
}
\caption{Set-level evaluation over 300 open-ended image editing tasks. Values report mean $\pm$ standard error across tasks. Higher is better for all metrics.}
\label{tab:main_results}
\end{table}

\textbf{Exploration is more effective in instruction space than through image-model stochasticity.} Across 300 tasks, instruction-space exploration produced broader and less redundant candidate sets than stochastic image sampling. With the same Qwen backend, direct image sampling covered 2.46 directions and produced 3.63 distinct realizations; Naive instruction sampling increased these to 2.90 and 4.42, while Realization only reached 4.17 and 5.63. This indicates that diversity is more effectively introduced before image synthesis than through image-model randomness alone.

\textbf{Image backends fail through reconstruction or repetition.} Gemini achieved the highest instruction-following score (3.93), but qualitative inspection revealed frequent structural hallucinations, such as changes to walls, architectural openings, room geometry, or viewpoint, as illustrated by the examples in Figure\ref{fig:qual_pic}. Its polished outputs therefore did not always remain grounded edits of the source scene. Flux exhibited the opposite failure mode: despite achieving the highest groundedness (3.95), its 5.71 valid outputs collapsed to only 2.85 distinct realizations across 1.49 directions. Strong individual outputs therefore do not necessarily provide useful set-level exploration.

\textbf{Realization Stage 2 drives diversity, while Region-level Stage 1 strengthens grounding.} Realization only already achieved the highest Direction Coverage (4.17) and produced 5.63 distinct realizations, showing that Stage 2 drives most of the candidate-set diversity. Adding a High-level exploration stage 1 produced nearly identical results, suggesting that an abstract semantic scaffold provides little additional benefit. Region-level Stage 1 instead achieved higher groundedness and Valid@8 while maintaining comparable realization diversity. Thus, Stage 1 is not inherently beneficial; when structured around feasible regions and edit extents, it primarily grounds the diverse realizations produced by Stage 2.

\section{Discussion}

Our results suggest that open-ended generation should be viewed not only as producing diverse samples, but as covering multiple valid modes of a constrained solution space. This is particularly important for image editing, where outputs must remain grounded in a specific source image. Simply increasing variation is therefore insufficient: repeated sampling may remain concentrated around similar high-level realizations, while generic diversity prompting can introduce less grounded alternatives. The goal is instead to broaden semantic coverage while remaining within the set of edits supported by the image and user request.

More broadly, this highlights a complementary perspective to the common goal of sharpening a model distribution toward a preferred output. For inherently open-ended tasks, multiple distinct modes may all be desirable, making the problem one of preserving and exposing valid multimodality rather than collapsing toward a single solution. Our current approach addresses this at inference time by structuring what can vary before generation; a longer-term question is how models can learn and represent such distributions over valid solutions directly.

\section{Conclusion}

We study open-ended image editing as a grounded exploration problem, where a useful system should expose multiple meaningfully different solutions without drifting from the source image or user request. Our results show that generic diversity prompting is not sufficient, while structuring the space of permissible variation can improve the balance between grounding and exploration. More broadly, this suggests that open-ended multimodal tasks may benefit from generation strategies that preserve and expose multiple valid modes rather than optimizing toward a single preferred realization.
\bibliographystyle{plainnat}
\bibliography{custom}
\appendix

\section{Additional Qualitative Results}
Due to space constraints, we show the first four of eight outputs from each method. In the source image, the left workspace and living area occupy different depth planes. Gemini consistently removes this boundary and reconstructs the two areas as a single continuous space. Region-Level Exploration
instead introduces varied edits while better preserving the original scene geometry. Prompted Diversity produces varied alternatives, but some involve larger structural changes or viewpoint shifts.
\begin{figure}[!htbp]
    \centering
    \includegraphics[width=\linewidth]{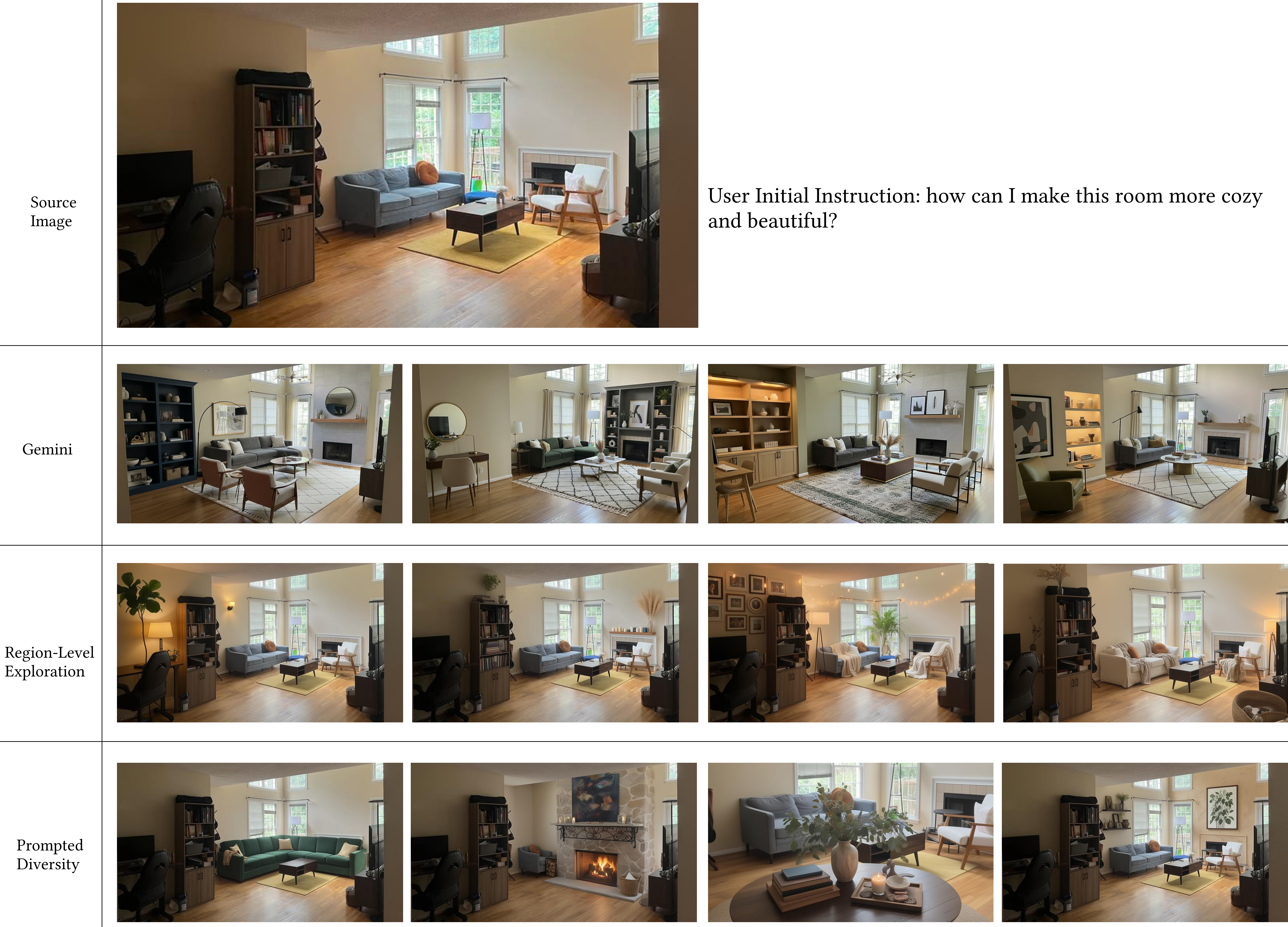}
    \caption{Qualitative comparison over Gemini, prompt diversity and region-level exploration}
    \label{fig:qual_pic}
\end{figure}


\definecolor{promptgray}{RGB}{248,248,248}
\definecolor{promptborder}{RGB}{210,210,210}

\lstdefinestyle{promptstyle}{
  basicstyle=\ttfamily\small,
  breaklines=true,
  breakatwhitespace=false,
  columns=fullflexible,
  keepspaces=true,
  showstringspaces=false,
  frame=single,
  rulecolor=\color{promptborder},
  backgroundcolor=\color{promptgray},
  framesep=6pt,
  xleftmargin=0pt,
  xrightmargin=0pt,
  aboveskip=6pt,
  belowskip=6pt
}

\section{Condensed Generation Prompts}
\label{app:generation-prompts}

All generation prompts receive the source image and original user request. We set all candidate counts to eight. Repeated input and output boilerplate is omitted for brevity.

\subsection{Realization Generation}

\begin{lstlisting}[style=promptstyle, numbers=none, breaklines=true, breakatwhitespace=true, columns=fullflexible]
Given the image, request, and optional Stage-1 scaffold, generate K concrete and executable edit instructions.

Each instruction must:
- plausibly satisfy the request;
- specify which visible content changes and how;
- use an appropriate edit extent;
- preserve source geometry and spatial relationships;
- preserve unrelated content and explicit constraints.

The set should differ in dominant editing direction or composition, not merely in wording or superficial attributes.

Return only a JSON array of K instruction strings.
\end{lstlisting}

\subsection{Region-Level Exploration}

\begin{lstlisting}[style=promptstyle, numbers=none, breaklines=true, breakatwhitespace=true, columns=fullflexible]
Construct K pairwise-different spatial edit scopes for the request.

Vary the amount and distribution of editable content. For each scope, identify the editable regions and a clear preservation boundary.

These are WHERE and EXTENT choices, not choices of style, material, object, palette, or treatment. Do not broaden a narrowly targeted request to unrelated regions.
\end{lstlisting}

\section{Condensed Evaluation Prompts}
\label{app:evaluation-prompts}

We use three structured GPT-5.5 judge calls. The prompts below omit repeated image-indexing instructions and diagnostic details while retaining the core evaluation criteria.

\subsection{Validity and Set-Level Ratings}

\begin{lstlisting}[style=promptstyle, numbers=none, breaklines=true, breakatwhitespace=true, columns=fullflexible]
Evaluate each candidate against the source image and original user request. Do not evaluate diversity.

A candidate is valid only if it:
- reasonably satisfies the requested intent and scope;
- remains an edit of the same source scene;
- preserves viewpoint, fixed structure, geometry, spatial relationships, and unrelated content unless their modification is required by the request;
- remains physically plausible and free of severe artifacts.

Unsupported reconstruction, structural or layout changes, unrelated edits, wrong targets, and required-edit no-ops are invalid.

Also rate the complete set:

Instruction Following:
5 = almost all outputs satisfy the request;
3 = mixed performance;
1 = most outputs fail.

Groundedness:
5 = source structure is preserved almost throughout;
3 = noticeable drift occurs in several outputs;
1 = most outputs depart from the source.

Return structured JSON containing both ratings and the validity decision for every candidate.
\end{lstlisting}

\subsection{High-Level Direction Coverage}

\begin{lstlisting}[style=promptstyle, numbers=none, breaklines=true, breakatwhitespace=true, columns=fullflexible]
Candidate validity has already been fixed. Partition all candidates into high-level semantic editing directions.

Use different directions for differences in editing target, function, layout, intervention strategy, or interpretation of the request.

Use the same direction when candidates differ only in color, material, texture, accessory choice, minor placement, or other low-level realization details.

Each candidate must appear exactly once. A direction is valid if it contains at least one valid candidate.

Return the direction partition as structured JSON.
\end{lstlisting}

\subsection{Distinct Realizations and Dispersion}

\begin{lstlisting}[style=promptstyle, numbers=none, breaklines=true, breakatwhitespace=true, columns=fullflexible]
All provided candidates have already passed the validity and grounding checks.

First, cluster candidates into distinct realizations. Keep candidates together when they express the same consequential edit decision and differ only in incidental appearance or stochastic rendering.

Split candidates for meaningfully different targets, operations, functions, layouts, furnishings, or decor solutions.

Then group the realization clusters into high-level directions using differences in target, function, layout, strategy, or interpretation.

Return both clustering levels as structured JSON.
\end{lstlisting}

\end{document}